\documentclass[runningheads]{llncs}
\usepackage{graphicx}
\usepackage{amsmath,amssymb} % define this before the line numbering.
\usepackage{color}
\newcommand{\red}[1]{\textcolor{red}{#1}}
\newcommand{\green}[1]{\textcolor{green}{#1}}
\usepackage{epstopdf}
\usepackage{booktabs}
\usepackage{subfig}
\usepackage[table]{xcolor}
\usepackage{diagbox}
\usepackage{caption}
\usepackage[labelfont=bf,labelsep=period]{caption}
\usepackage{lmodern}
\usepackage[hyphens]{url}
\usepackage{cite}

\begin{document}
\title{Zero-Shot Deep Domain Adaptation}
% Replace with your title

\titlerunning{Zero-Shot Deep Domain Adaptation}
% Replace with a meaningful short version of your title
%
\author{Kuan-Chuan Peng \and
Ziyan Wu \and
Jan Ernst}
%
%Please write out author names in full in the paper, i.e. full given and family names.
%If any authors have names that can be parsed into FirstName LastName in multiple ways, please include the correct parsing, in a comment to the volume editors:
%\index{Lastnames, Firstnames}
%(Do not uncomment it, because you may introduce extra index items if you do that, we will use scripts for introducing index entries...)
\authorrunning{K.-C. Peng, Z. Wu, and J. Ernst}
% Replace with shorter version of the author list. If there are more authors than fits a line, please use A. Author et al.
%

\institute{Siemens Corporate Technology, Princeton NJ 08540, USA\\
\email{\{kuanchuan.peng, ziyan.wu, jan.ernst\}@siemens.com}}
\maketitle              % typeset the header of the contribution
\begin{abstract}
Domain adaptation is an important tool to transfer knowledge about a task (e.g. classification) learned in a source domain to a second, or target domain. Current approaches assume that task-relevant target-domain data is available during training. We demonstrate how to perform domain adaptation when no such task-relevant target-domain data is available. To tackle this issue, we propose \emph{zero-shot deep domain adaptation} (ZDDA), which uses privileged information from \emph{task-irrelevant dual-domain pairs}. ZDDA learns a source-domain representation which is not only tailored for the task of interest but also close to the target-domain representation. Therefore, the source-domain task of interest solution (e.g. a classifier for classification tasks) which is jointly trained with the source-domain representation can be applicable to both the source and target representations. Using the MNIST, Fashion-MNIST, NIST, EMNIST, and SUN RGB-D datasets, we show that ZDDA can perform domain adaptation in classification tasks without access to task-relevant target-domain training data. We also extend ZDDA to perform sensor fusion in the SUN RGB-D scene classification task by simulating task-relevant target-domain representations with task-relevant source-domain data. To the best of our knowledge, ZDDA is the first domain adaptation and sensor fusion method which requires no task-relevant target-domain data. The underlying principle is not particular to computer vision data, but should be extensible to other domains.

\keywords{zero-shot \and domain adaptation \and sensor fusion}
\end{abstract}
\section{Introduction}\label{sec:intro}
The useful information to solve practical tasks often exists in different domains captured by various sensors, where a domain can be either a modality or a dataset. For instance, the 3-D layout of a room can be either captured by a depth sensor or inferred from the RGB images. In real-world scenarios, it is highly likely that we can only access limited amount of data in certain domain(s). The performance of the solution (e.g. the classifier for classification tasks) we learn from one domain often degrades when the same solution is applied to other domains, which is caused by domain shift~\cite{GrettonMITPress09} in a typical domain adaptation (DA) task, where source-domain training data, target-domain training data, and a task of interest (TOI) are given. The goal of a DA task is to derive solution(s) of the TOI for both the source and target domains.

\begin{figure}[t]
    \centering
    \includegraphics[width=0.75\linewidth]{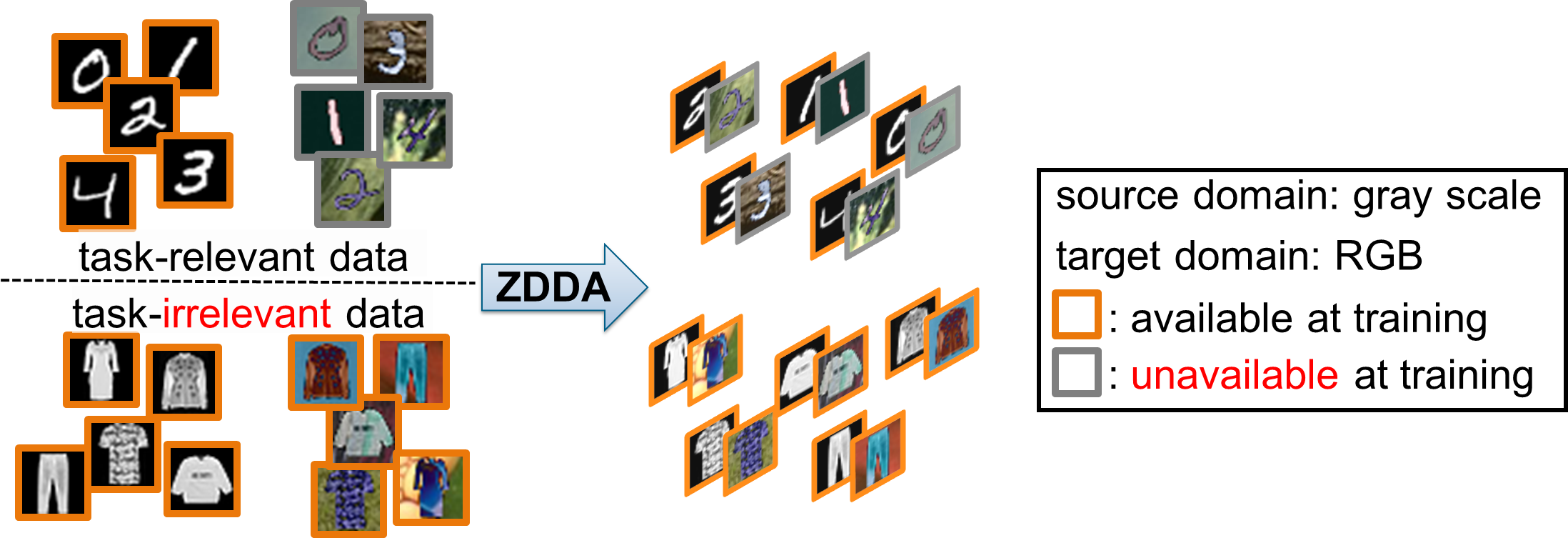}
    \caption{\fontsize{9}{10.8} \selectfont We propose zero-shot deep domain adaptation (ZDDA) for domain adaptation and sensor fusion. ZDDA learns from the task-irrelevant dual-domain pairs when the task-relevant target-domain training data is unavailable. In this example domain adaptation task (MNIST~\cite{LeCunIEEE98}$\rightarrow$MNIST-M~\cite{GaninICML15}), the task-irrelevant gray-RGB pairs are from the Fashion-MNIST~\cite{XiaoCoRR17} dataset and the Fashion-MNIST-M dataset (the colored version of the Fashion-MNIST~\cite{XiaoCoRR17} dataset with the details in Sec.~\ref{subsec:dataset})}
    \label{idea}
\end{figure}

The state-of-the-art DA methods such as~\cite{AljundiECCVW16,GaninJMLR16,GebruICCV17,GhifaryECCV16,KoniuszCVPR17,MotiianICCV17,SohnICCV17,SunECCVW16,TzengICCV15,TzengCVPR17,VenkateswaraCVPR17,WangICCV17,WuCVPR17,YanCVPR17,ZhangCVPR17} are proposed to solve DA tasks under the assumption that the \textit{task-relevant data}, the data directly applicable and related to the TOI (regardless of whether it is labeled or not), in the target domain is available at training time, which is not always true in practice. For instance, in real business use cases, acquiring the task-relevant target-domain training data can be infeasible due to the combination of the following reasons: 1) Unsuitable tools at the field. 2) Product development timeline. 3) Budget limitation. 4) Data import/export regulations. Such impractical assumption is also assumed true in the existing works of sensor fusion such as~\cite{NgiamICML11,YangCVPR17}, where the goal is to obtain a dual-domain (source and target) TOI solution which is robust to noise in either domain. This unsolved issue motivates us to propose \textit{zero-shot deep domain adaptation} (ZDDA), a DA and sensor fusion approach which learns from the task-irrelevant dual-domain training pairs without using the task-relevant target-domain training data, where we use the term \textit{task-irrelevant data} to refer to the data which is not task-relevant. In the rest of the paper, we use T-R and T-I as the shorthand of task-relevant and task-irrelevant, respectively.

We illustrate what ZDDA is designed to achieve in Fig.~\ref{idea} using an example DA task (MNIST~\cite{LeCunIEEE98}$\rightarrow$MNIST-M~\cite{GaninICML15}). We recommend that the readers view all the figures and tables in color. In Fig.~\ref{idea}, the source and target domains are gray scale and RGB images respectively, and the TOI is digit classification with both the MNIST~\cite{LeCunIEEE98} and MNIST-M~\cite{GaninICML15} testing data. We assume that the MNIST-M~\cite{GaninICML15} training data is unavailable. In this example, ZDDA aims at using the MNIST~\cite{LeCunIEEE98} training data and the T-I gray-RGB pairs from the Fashion-MNIST~\cite{XiaoCoRR17} dataset and the Fashion-MNIST-M dataset (the colored version of the Fashion-MNIST~\cite{XiaoCoRR17} dataset with the details in Sec.~\ref{subsec:dataset}) to train digit classifiers for MNIST~\cite{LeCunIEEE98} and MNIST-M~\cite{GaninICML15} images. Specifically, ZDDA achieves this by simulating the RGB representation using the gray scale image and building a joint network with the supervision of the TOI in the gray scale domain. We present the details of ZDDA in Sec.~\ref{sec:method}.

We make the following two contributions: (1) To the best of our knowledge, our proposed method, ZDDA, is the first deep learning based method performing domain adaptation between one source image modality and another different target image modality (not just different datasets in the same modality such as the Office dataset~\cite{SaenkoECCV10}) \textbf{without using the task-relevant target-domain training data}. We show ZDDA's efficacy using the MNIST~\cite{LeCunIEEE98}, Fashion-MNIST~\cite{XiaoCoRR17}, NIST~\cite{Grother16}, EMNIST~\cite{CohenCoRR17}, and SUN RGB-D~\cite{SongCVPR15} datasets with cross validation. (2) \textbf{Given no task-relevant target-domain training data}, we show that ZDDA can perform sensor fusion and that ZDDA is more robust to noisy testing data in either source or target or both domains compared with a naive fusion approach in the scene classification task from the SUN RGB-D~\cite{SongCVPR15} dataset.

\section{Related Work}

Domain adaptation (DA) has been extensively studied in computer vision and applied to various applications such as image classification~\cite{AljundiECCVW16,GaninJMLR16,GebruICCV17,GhifaryECCV16,KoniuszCVPR17,MotiianICCV17,SohnICCV17,SunECCVW16,TzengICCV15,TzengCVPR17,VenkateswaraCVPR17,WangICCV17,WuCVPR17,YanCVPR17,ZhangCVPR17}, semantic segmentation~\cite{WulfmeierIROS17,ZhangICCV17}, and image captioning~\cite{ChenICCV17}. With the advance of deep neural networks in recent years, the state-of-the-art methods successfully perform DA with (fully or partially) labeled~\cite{ChenICCV17,GebruICCV17,KoniuszCVPR17,MotiianICCV17,TzengICCV15} or unlabeled~\cite{AljundiECCVW16,GaninJMLR16,GebruICCV17,GhifaryECCV16,SohnICCV17,SunECCVW16,TzengICCV15,TzengCVPR17,VenkateswaraCVPR17,WangICCV17,WuCVPR17,WulfmeierIROS17,YanCVPR17,ZhangCVPR17} T-R target-domain data. Although different strategies such as the domain adversarial loss~\cite{TzengCVPR17} and the domain confusion loss~\cite{TzengICCV15} are proposed to improve the performance in the DA tasks, most of the existing methods need the T-R target-domain training data, which can be unavailable in reality. In contrast, we propose ZDDA to learn from the T-I dual-domain pairs without using the T-R target-domain training data. One part of ZDDA includes simulating the target-domain representation using the source-domain data, and similar concepts have been mentioned in~\cite{GuptaCVPR16,HoffmanCVPR16}. However, both of~\cite{GuptaCVPR16,HoffmanCVPR16} require the access to the T-R dual-domain training pairs, but ZDDA needs no T-R target-domain data.

\begin{table}[t]
\centering
\caption{\fontsize{9}{10.8} \selectfont Problem setting comparison between ZDDA, unsupervised domain adaptation (UDA), multi-view learning (MVL), and domain generalization (DG)
}
\begin{tabular}{ccccc}
\\
\toprule
problem conditions &UDA &MVL &DG &ZDDA\\
\midrule
given T-R target-domain training data? &\green{Y} &\green{Y} &\red{N} &\red{N}\\
given T-R training data in multiple ($>$1) domains/views? &\red{N} &\green{Y} &\green{Y} &\red{N}\\
example prior work &\cite{SaitoICML17} &\cite{WangICML15} &\cite{LiICCV17} &N/A\\
\bottomrule
\end{tabular}
\label{compare_problem}
\end{table}

\begin{table}[t]
\centering
\caption{\fontsize{9}{10.8} \selectfont Working condition comparison between ZDDA and other existing methods. Among all the listed methods, only ZDDA can work under all four conditions
}
%\fontsize{8}{9.6} \selectfont

\begin{tabular}{c@{\hspace{0mm}}c@{\hspace{0mm}}c@{\hspace{0mm}}c@{\hspace{0mm}}c@{\hspace{0mm}}c@{\hspace{0mm}}c@{\hspace{0mm}}c@{\hspace{0.5mm}}c}
\\
\toprule
Can each method work under each condition? &\cite{LiICCV17} &\cite{DingTIP15} &\cite{Blitzer09} &\cite{YangBMVCW15} &\cite{GaninJMLR16} &\cite{SaenkoECCV10} &\cite{TzengICCV15} &ZDDA\\
\midrule
\textbf{without} T-R target-domain training data &\green{Y} &\red{N} &\red{N} &\green{Y} &\red{N} &\green{Y} &\red{N} &\green{Y}\\
\textbf{without} T-R training data in $>$1 domains &\red{N} &\green{Y} &\green{Y} &\green{Y} &\green{Y} &\green{Y} &\green{Y} &\green{Y}\\
\textbf{without} accurate domain descriptor &\green{Y} &\green{Y} &\green{Y} &\red{N} &\green{Y} &\green{Y} &\green{Y} &\green{Y}\\
\textbf{without} class labels for \textbf{any} target domain data &\green{Y} &\red{N} &\green{Y} &\green{Y} &\green{Y} &\red{N} &\red{N} &\green{Y}\\
\midrule
\cellcolor{blue!30}conjunction of all the above conditions &\cellcolor{blue!30}\textbf{\red{N}} &\cellcolor{blue!30}\textbf{\red{N}} &\cellcolor{blue!30}\textbf{\red{N}} &\cellcolor{blue!30}\textbf{\red{N}} &\cellcolor{blue!30}\textbf{\red{N}} &\cellcolor{blue!30}\textbf{\red{N}} &\cellcolor{blue!30}\textbf{\red{N}} &\cellcolor{blue!30}\textbf{\green{Y}}\\
\bottomrule
\end{tabular}
\label{compare_problem_rebuttal}
\end{table}

Other problems related to ZDDA include unsupervised domain adaptation (UDA), multi-view learning (MVL), and domain generalization (DG), and we compare their problem settings in Table~\ref{compare_problem}, which shows that the ZDDA problem setting is different from those of UDA, MVL, and DG. In UDA and MVL, T-R target-domain training data is given. In MVL and DG, T-R training data in multiple domains is given. However, in ZDDA, T-R target-domain training data is unavailable and the only available T-R training data is in one source domain. We further compare ZDDA with the existing methods relevant to our problem setting in Table~\ref{compare_problem_rebuttal}, which shows that among the listed methods, only ZDDA can work under all four conditions.

%Ding et al.~\cite{DingTIP15} proposed a missing modality transfer learning problem which has a closely related problem setting as that of ZDDA. However, in~\cite{DingTIP15}, the data in the source and target domains of the auxiliary and objective databases is semantically identical (they are all faces), but in ZDDA, the auxiliary and objective databases are T-I and T-R data respectively (with high semantic irrelevance with respect to each other; for example, digits vs. clothing in our experiment). Therefore,~\cite{DingTIP15} still have access to the T-R target-domain data, which is not the case for ZDDA.
%
%The name ``zero-shot domain adaptation" has already been used by Blitzer et al.~\cite{Blitzer09} and Yang et al.~\cite{YangBMVCW15}. However, both methods require additional prerequisites. Blitzer's method~\cite{Blitzer09} needs access to the T-R target-domain data. Requiring that the domain descriptor is always observed and accurate, Yang's method~\cite{YangBMVCW15} shows its efficacy using only different datasets in the same modality (specifically, the Office dataset~\cite{SaenkoECCV10}) instead of the data in different modalities. ZDDA requires none of these two prerequisites, and we show the efficacy of ZDDA with the data in different modalities.

In terms of sensor fusion, Ngiam et al.~\cite{NgiamICML11} define the three components for multimodal learning (multimodal fusion, cross modality learning, and shared representation learning) based on the modality used for feature learning, supervised training, and testing, and experiment on audio-video data with their proposed deep belief network and autoencoder based method. Targeting on the temporal data, Yang et al.~\cite{YangCVPR17} follow the setup of multimodal learning in~\cite{NgiamICML11}, and validate their proposed encoder-decoder architecture using video-sensor and audio-video data. Although certain progress about sensor fusion is achieved in the previous works~\cite{NgiamICML11,YangCVPR17}, we are unaware of any existing sensor fusion method which overcomes the issue of lacking T-R target-domain training data, which is the issue that ZDDA is designed to solve.

\section{Our Proposed Method --- ZDDA}\label{sec:method}

%\begin{figure*}
%    \centering
%    \subfloat[\normalsize an overview of our training procedure]{%
%        \includegraphics[width=1\linewidth]{image/training_flow.png}%
%        \label{training_flow}%
%        }%
%    \\
%    %\hfill%
%    \subfloat[\normalsize an illustration of the motivation of (a)]{%
%        \includegraphics[width=.7\linewidth]{image/training_concept.png}%
%        \label{training_flow_concept}%
%        }%
%    \caption{We use the images from the SUN RGB-D~\cite{SongCVPR15} dataset for illustration in Figure~\ref{training_flow}, where we simulate the target-domain representation using the source-domain data, build a joint network with the supervision from the source domain, and train a domain fusion network. We explain the concept behind Figure~\ref{training_flow} in Figure~\ref{training_flow_concept}, where $f_x$ denotes the features learned by $CNN_x$, and $f_{TOI}$ is the ideal features for our TOI. The details are explained in Sec.~\ref{sec:method}.}
%    \label{training_flow_explanation}
%\end{figure*}

Given a task of interest (TOI), a source domain $D_s$, and a target domain $D_t$, our proposed method, zero-shot deep domain adaptation (ZDDA), is designed to achieve the following two goals: 1) \textbf{Domain adaptation}: Derive the solutions of the TOI for both $D_s$ and $D_t$ when the T-R training data in $D_t$ is unavailable. We assume that we have access to the T-R labeled training data in $D_s$ and the T-I dual-domain pairs in $D_s$ and $D_t$. 2) \textbf{Sensor fusion}: Given the previous assumption, derive the solution of TOI when the testing data in both $D_s$ and $D_t$ is available. The testing data in either $D_s$ or $D_t$ can be noisy. We assume that there is no prior knowledge available about the type of noise and which domain gives noisy data at testing time.

\begin{figure}[t]
\centering
    \includegraphics[width=1\linewidth]{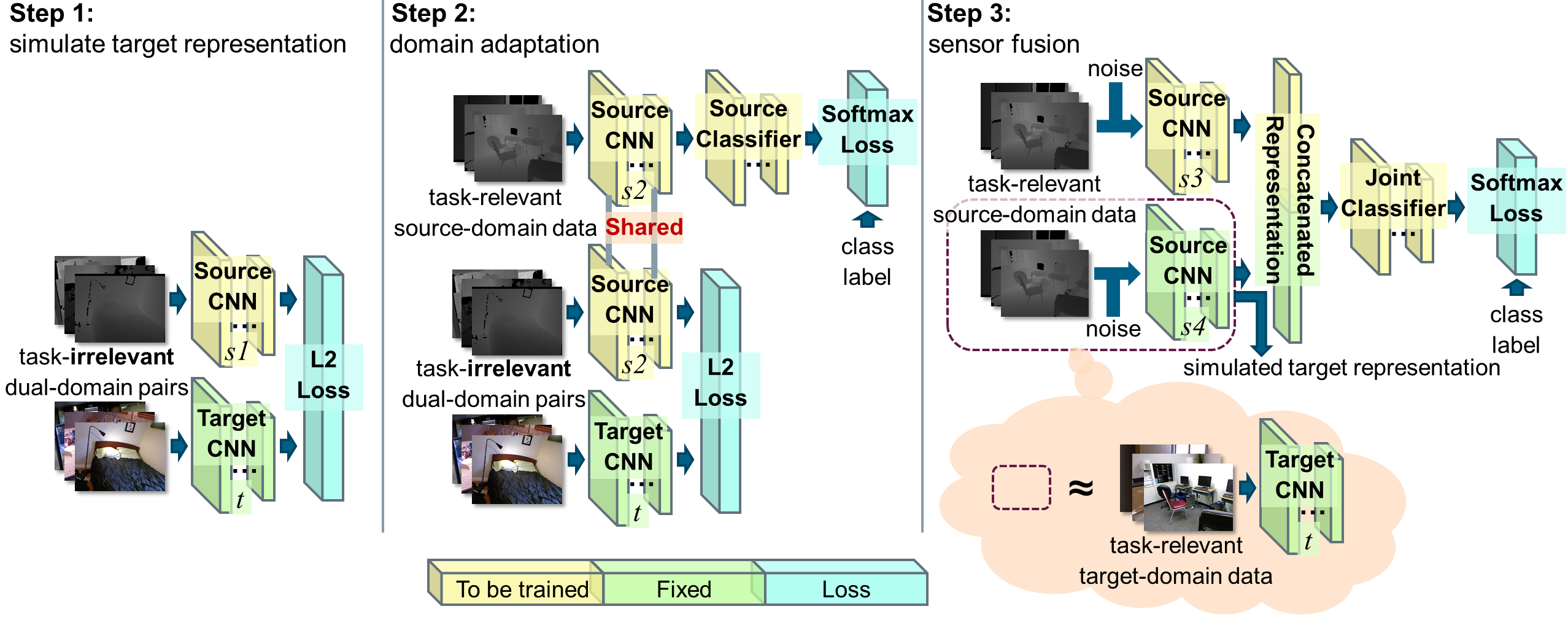}
    \caption{\fontsize{9}{10.8} \selectfont An overview of the ZDDA training procedure. We use the images from the SUN RGB-D~\cite{SongCVPR15} dataset for illustration. ZDDA simulates the target-domain representation using the source-domain data, builds a joint network with the supervision from the source domain, and trains a sensor fusion network. In step 1, we choose to train $s$\textit{1} and fix $t$, but we can also train $t$ and fix $s$\textit{1} to simulate the target-domain representation. In step 2, $t$ can also be trainable instead of being fixed, but we choose to fix it to make the number of trainable parameters manageable. The details are explained in Sec.~\ref{sec:method}}
    \label{training_flow}
\end{figure}

For convenience, we use a scene classification task in RGB-D as an example TOI to explain ZDDA, but ZDDA can be applied to other TOIs/domains. In this example, $D_s$ and $D_t$ are depth and RGB images respectively. According to the our previous assumption, we have access to the T-R labeled depth data and T-I RGB-D pairs at training time. The training procedure of ZDDA is illustrated in Fig.~\ref{training_flow}, where we simulate the RGB representation using the depth image, build a joint network with the supervision of the TOI in depth images, and train a sensor fusion network in step 1, step 2, and step 3 respectively. We use the ID marked at the bottom of each convolutional neural networks (CNN) in Fig.~\ref{training_flow} to refer to each CNN.

In step 1, we create two CNNs, $s$\textit{1} and $t$, to take the depth and RGB images of the T-I RGB-D pairs as input. The purpose of this step is to find $s$\textit{1} and $t$ such that feeding the RGB image into $t$ can be approximated by feeding the corresponding depth image into $s$\textit{1}. We achieve this by fixing $t$ and enforcing the L2 loss on top of $s$\textit{1} and $t$ at training time. We choose to train $s$\textit{1} and fix $t$ here, but training $t$ and fixing $s$\textit{1} can also achieve the same purpose. The L2 loss can be replaced with any suitable loss functions which encourage the similarity of the two input representations, and our selection is inspired by~\cite{GuptaCVPR16,HoffmanCVPR16}. The design in step 1 is similar to the hallucination architecture~\cite{HoffmanCVPR16} and the supervision transfer~\cite{GuptaCVPR16}, but we require no T-R dual-domain training pairs. Instead, we use the T-I dual-domain training pairs.

After step 1, we add another CNN, $s$\textit{2} (with the same network architecture as that of $s$\textit{1}), and a classifier to the network (as shown in step 2) to learn from the label of the training depth images. The classifier in our experiment is a fully connected layer for simplicity, but other types of classifiers can also be used. The newly added CNN takes the T-R depth images as input, and shares all the weights with the original source CNN, so we use $s$\textit{2} to refer to both of them. $t$ is the same as that in step 1. At training time, we pre-train $s$\textit{2} from $s$\textit{1} and fix $t$. Our choice of fixing $t$ is inspired by the adversarial adaptation step in ADDA~\cite{TzengCVPR17}. $t$ can also be trainable in step 2, but given our limited amount of data, we choose to fix it to make the number of trainable parameters manageable. $s$\textit{2} and the source classifier are trained such that the weighted sum of the softmax loss and L2 loss are minimized. The softmax loss can be replaced with other losses suitable for the TOI.

\begin{figure}[t]
    \centering
    \subfloat[\small testing domain adaptation]{%
        \includegraphics[height=3.5cm]{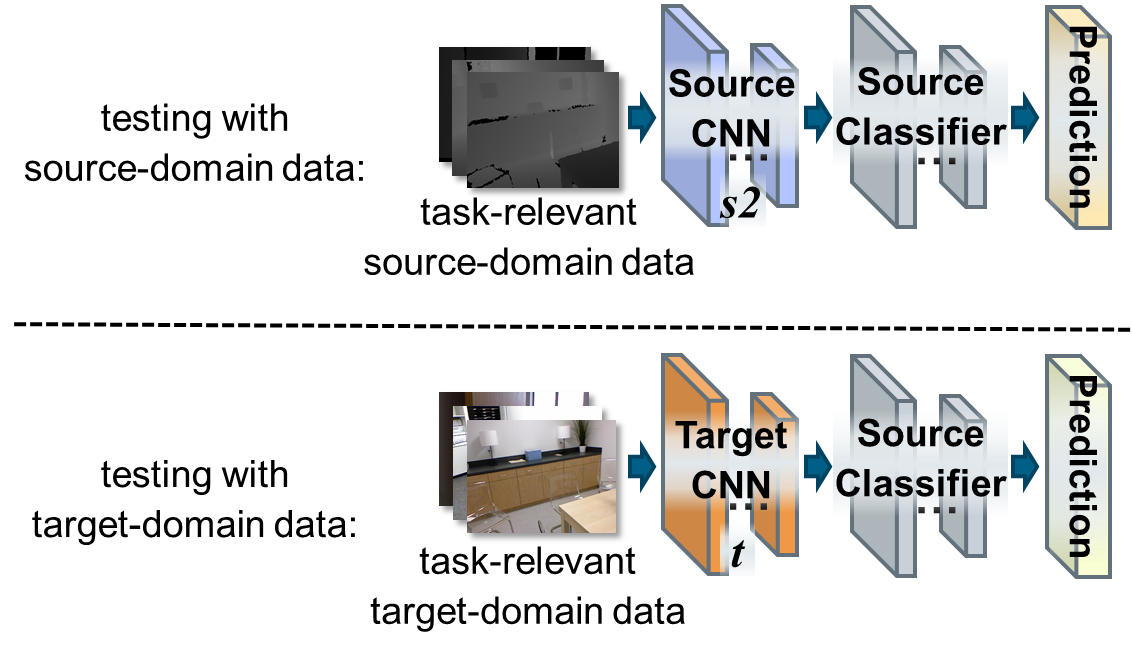}%
        \label{testing_flow:a}%
        }%
    \subfloat[\small testing sensor fusion]{%
        \includegraphics[height=3.5cm]{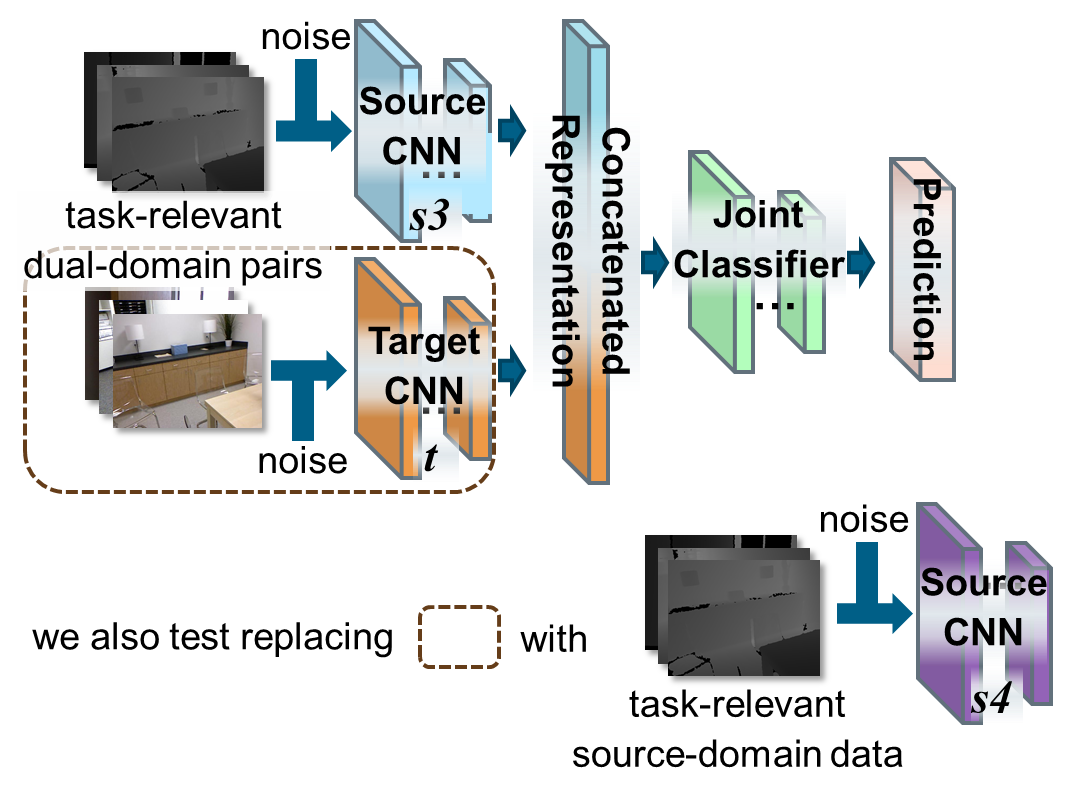}%
        \label{testing_flow:b}%
        }%
    \caption{\fontsize{9}{10.8} \selectfont An overview of the ZDDA testing procedure. We use the SUN RGB-D~\cite{SongCVPR15} images for illustration. Different from the color coding in Fig.~\ref{training_flow}, the colors here are purely used to distinguish different CNNs/classifiers/predictions}
    \label{testing_flow}
\end{figure}

After step 2, we expect to obtain a depth representation which is close to the RGB representation in the feature space and performs reasonably well with the trained classifier in the scene classification. Step 1 and step 2 can be done in one step with properly designed curriculum learning, but we separate them not only because of clarity but also because of the difficulty of designing the learning curriculum before training. After step 2, we can form the scene classifier in depth/RGB (denoted as $C_D$/$C_{RGB}$) by concatenating $s$\textit{2}/$t$ and the trained source classifier (as shown in Fig.~\ref{testing_flow:a}), which meets our first goal, domain adaptation. We use the notation ZDDA$_2$ to refer to the method using the training procedure in Fig.~\ref{training_flow} up to step 2 and the testing procedure in Fig.~\ref{testing_flow:a}.

To perform sensor fusion, we propose step 3, where we train a joint classifier for RGB-D input using only the T-R depth training data. We create two CNNs, $s$\textit{3} and $s$\textit{4} (each with the same network architecture as that of $CNN_{s\text{1}}$), and add a concatenation layer on top of them to concatenate their output representations. The concatenated representation is connected to a joint classifier. At training time, we pre-train $s$\textit{3} and $s$\textit{4} from $s$\textit{2} and $s$\textit{1} respectively and fix $s$\textit{4}. Both $s$\textit{3} and $s$\textit{4} take the T-R depth images as the input. To train a more robust RGB-D scene classifier, we randomly select some inputs of $s$\textit{3} and $s$\textit{4}, and optionally add noise to them independently. We supervise the entire network with the label of the depth training data for the scene classification, which is done by the softmax loss enforced on top of the joint classifier.

According to step 1, the output of $s$\textit{4} is expected to simulate the RGB representation as if we feed the T-R RGB image to $t$. This expectation is based on the assumption that the relationship between the dual-domain pairwise data is similar, regardless of whether the data is T-R or T-I. Given the simulated RGB representation, $s$\textit{3} is trained to learn a depth representation suitable for the RGB-D scene classification without the constraint of the L2 loss in step 2. At testing time, $s$\textit{4} is replaced with $t$ which takes the T-R RGB testing images as input with optional noise added to test the ZDDA's performance given noisy RGB-D testing data (as shown in Fig.~\ref{testing_flow:b}). In Fig.~\ref{testing_flow:b}, we also test replacing ``RGB images and $t$" with ``depth images and $s$\textit{4}" to evaluate the performance of ZDDA in step 3 given only testing depth images. After the training procedure in Fig.~\ref{training_flow}, we can form three scene classifiers in RGB, depth, and RGB-D domains (one classifier per domain), and our trained RGB-D classifier is expected to be able to handle noisy input with reasonable performance degradation. The 3-step training procedure of ZDDA in Fig.~\ref{training_flow} can be framed as an end-to-end training process with proper learning curriculum. We separate these 3 steps due to the ease of explanation. We use the notation ZDDA$_3$ to refer to the method using the training procedure in Fig.~\ref{training_flow} up to step 3 and the testing procedure in Fig.~\ref{testing_flow:b}.

\section{Experiment Setup}\label{sec:exp}

\subsection{Datasets}\label{subsec:dataset}

\begin{table}[t]
\centering
\fontsize{8}{9.6} \selectfont
\caption{
\fontsize{9}{10.8} \selectfont The statistics of the datasets we use. For NIST, we use the ``by\_class" dataset, remove the digits, and treat uppercase and lowercase letters as different classes. For EMNIST, we use the ``EMNIST Letters" split which only contains the letters. We create the colored datasets from the original ones using Ganin's method~\cite{GaninICML15} (see Sec.~\ref{subsec:dataset} for details). We refer to each dataset by the corresponding dataset ID (e.g. $D_N$ and $D_N$-M refer to the NIST and the NIST-M datasets, respectively)
}
\begin{tabular}{cccccc}
\\
\toprule
original dataset &MNIST~\cite{LeCunIEEE98} &Fashion-MNIST~\cite{XiaoCoRR17} &NIST~\cite{Grother16} &EMNIST~\cite{CohenCoRR17} &SUN RGB-D~\cite{SongCVPR15}\\
\hline
dataset ID &$D_M$ &$D_F$ &$D_N$ &$D_E$ &$D_S$\\
\hline
image content &digit &clothing &letter &letter &scene\\
\midrule
image size &28$\times$28 &28$\times$28 &128$\times$128 &28$\times$28 &$\sim$VGA\\
\# classes &10 &10 &52 &26 &45\\
\# training data &60000 &60000 &387361 &124800 &details in Sec.~\ref{subsec:dataset}\\
\# testing data &10000 &10000 &23941 &20800 &details in Sec.~\ref{subsec:dataset}\\
class labels &0-9 &dress, coat, etc. &A-Z, a-z &Aa-Zz &corridor, lab, etc.\\
balanced class? &N &Y &N &Y &N\\
\raisebox{0.5\height}{example images} &\includegraphics[width=0.1\linewidth]{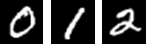} &\includegraphics[width=0.1\linewidth]{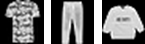} &\includegraphics[width=0.1\linewidth]{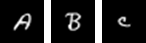} &\includegraphics[width=0.1\linewidth]{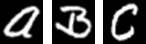} &\includegraphics[width=0.17\linewidth]{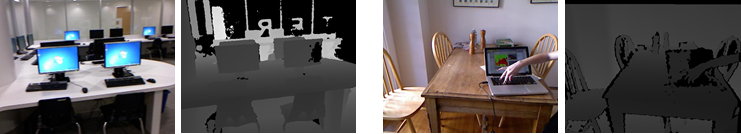}\\
\midrule[.75pt]
colored dataset &MNIST-M &Fashion-MNIST-M &NIST-M &EMNIST-M &N / A\\
\raisebox{0.5\height}{example images} &\centering\includegraphics[width=0.1\linewidth]{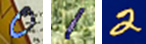} &\includegraphics[width=0.1\linewidth]{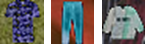} &\includegraphics[width=0.1\linewidth]{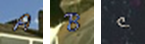} &\includegraphics[width=0.1\linewidth]{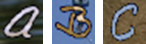} &\raisebox{0.5\height}{N / A}\\
\bottomrule
\end{tabular}
\label{dataset_stats}
\end{table}

For domain adaptation (DA), we validate the efficacy of ZDDA under classification tasks using the MNIST~\cite{LeCunIEEE98}, Fashion-MNIST~\cite{XiaoCoRR17}, NIST~\cite{Grother16}, EMNIST~\cite{CohenCoRR17}, and SUN RGB-D~\cite{SongCVPR15} datasets. For sensor fusion, we experiment on the SUN RGB-D~\cite{SongCVPR15} dataset. We summarize the statistics of these datasets in Table~\ref{dataset_stats}, where we list the dataset IDs which we use to refer to these datasets. For $D_M$, $D_F$, $D_N$, and $D_E$, we create the colored version of these datasets ($D_M$-M, $D_F$-M, $D_N$-M, and $D_E$-M) according to the procedure proposed in Ganin's work~\cite{GaninICML15} --- blending the gray scale images with the patches randomly extracted from the BSDS500 dataset~\cite{ArbelaezTPAMI11}. These colored datasets and the original ones are used to construct four DA tasks adapting from gray scale to RGB images. For each DA task, we use one of the other three pairs of the datasets (original and colored ones) as the T-I data. For example, for the DA task $D_M\rightarrow D_M$-M, $D_F$ and $D_F$-M together are one possible choice as the T-I data. The DA task $D_M\rightarrow D_M$-M is acknowledged as one of the standard experiments to test the efficacy of the DA methods in recent works~\cite{AljundiECCVW16,BousmalisCVPR17,GaninJMLR16,HaeusserICCV17,SaitoICML17,SenerNIPS16}, so we adopt this experiment and extend it to $D_F$, $D_N$, and $D_E$.

$D_S$ contains 10335 RGB-D pairs belonging to 45 different scenes. For each RGB-D pair, both the raw (noisy) depth image and post-processed clean depth image are provided, and we choose to use the raw depth image to simulate the real-world scenarios. Out of the 45 scenes, we select the following 10 scenes: computer room (0), conference room (1), corridor (2), dining room (3), discussion area (4), home office (5), idk (6), lab (7), lecture theatre (8), and study space (9), where the number after each scene is the scene ID we use to refer to each scene. The 8021 RGB-D pairs belonging to the other scenes are used as the T-I training data. The 10 scenes are selected based on the following two constraints: 1) Each scene contains at least 150 RGB-D pairs in $D_S$, which ensures a reasonable amount of T-R data. 2) The total number of the RGB-D pairs belonging to the selected 10 scenes is minimized, which maximizes the amount of the T-I training data. We empirically find that the amount and diversity of the T-I training data are important for ZDDA. To avoid the bias toward the scene with more data, for each of the selected 10 scenes, we randomly select 89/38 RGB-D pairs as the T-R training/testing data. When experimenting on different scene classification tasks using different selections of scenes, we only use the training/testing data associated with those selected scenes as the T-R data.

\subsection{Training Details}\label{subsec:training_details}

\begin{table}[t]
\centering
\caption{\fontsize{9}{10.8} \selectfont The base network architecture (BNA) we use in our experiments. For each BNA, We specify the layer separating the source/target CNN and the source classifier in Fig.~\ref{training_flow}. The layer name in the right column is based on the official Caffe~\cite{JiaCoRR14} and SqueezeNet\_v1.1~\cite{IandolaCoRR16} implementation of each BNA
}
\begin{tabular}{cc}
\\
\toprule
base network &source/target CNN architecture\\
architecture (BNA) &(up to where in BNA (inclusive))\\
\midrule
LeNet~\cite{LeNetPrototxt} &ip1\\
GoogleNet~\cite{SzegedyCVPR15} &pool5/7x7\_s1\\
AlexNet~\cite{KrizhevskyNIPS12} &fc7\\
SqueezeNet\_v1.1~\cite{IandolaCoRR16} &fire9/concat\\
\bottomrule
\end{tabular}
\label{NN_architecture}
\end{table}

We use Caffe~\cite{JiaCoRR14} to implement ZDDA. Table~\ref{NN_architecture} lists the base network architecture (BNA) we use and the layer separating the source/target CNN and the source classifier in Fig.~\ref{training_flow}. For instance, in the case when the BNA is LeNet~\cite{LeNetPrototxt}, the architecture of each source/target CNN in Fig.~\ref{training_flow} is the LeNet~\cite{LeNetPrototxt} architecture up to the ``ip1" layer, and the rest of the LeNet~\cite{LeNetPrototxt} architecture is used as the source classifier. For the DA tasks involving $D_M$, $D_F$, $D_N$, and $D_E$, we use the LeNet~\cite{LeNetPrototxt} as the BNA and train all the CNNs in Fig.~\ref{training_flow} from scratch except that the target CNN is pre-trained from the T-I dataset and fixed afterwards. For example, when using $D_F$ and $D_F$-M as the T-I data in the DA task $D_M\rightarrow D_M$-M, we train a CNN (denoted as $CNN_{ref}$) with the LeNet~\cite{LeNetPrototxt} architecture from scratch using the images and labels of $D_F$-M, and pre-train the target CNNs in Fig.~\ref{training_flow} from $CNN_{ref}$. We follow similar procedures for other DA tasks and T-I datasets involving $D_M$, $D_F$, $D_N$, and $D_E$.

For the experiment involving $D_S$, we mostly use GoogleNet~\cite{SzegedyCVPR15} as the BNA, but we also use AlexNet~\cite{KrizhevskyNIPS12} and SqueezeNet\_v1.1~\cite{IandolaCoRR16} in the cross validation experiment with respect to different BNAs. Since only limited amount of RGB-D pairs are available in $D_S$, we pre-train all the CNNs in Fig.~\ref{training_flow} from the BVLC GoogleNet model~\cite{BVLCGoogleNetModel}, BVLC AlexNet model~\cite{BVLCAlexNetModel}, and the reference SqueezeNet model~\cite{SqueezeNetModel} when the BNA is GoogleNet~\cite{SzegedyCVPR15}, AlexNet~\cite{KrizhevskyNIPS12}, and SqueezeNet\_v1.1~\cite{IandolaCoRR16}, respectively. These pre-trained models are trained for the ImageNet~\cite{DengCVPR09} classification task.

For the optionally added noise in ZDDA$_3$, we experiment on training/testing with noise-free data and noisy data. In the latter case, given that no prior knowledge about the noise is available, we use the black image as the noisy image to model the extreme case where no information in the noisy image is available. We train ZDDA$_3$ step 3 with the augmented training data formed by copying the original T-R source-domain training data 10 times and replacing $p_{train}\%$ of the images selected randomly with the black images. We follow this procedure twice independently and use the two augmented training datasets as the inputs of the two source CNNs in step 3. We empirically set $p_{train}=20$. The testing data in Fig.~\ref{testing_flow:b} is constructed by replacing $p_{test}\%$ of the original testing images selected randomly with the black images, and we evaluate ZDDA under different $p_{test}$s. For all the experiments, the number of the output nodes of the source/joint classifiers is set to be the number of classes in the TOI, and these classifiers are trained from scratch. For the joint classifiers, we use two fully connected layers unless otherwise specified, where the first fully connected layer of the joint classifier has 1024 output nodes.

%The visualization of the network architectures of the source CNN, source classifier, and joint classifier for each BNA is shown in the supplementary material.

In terms of the training parameters used in Fig.~\ref{training_flow} for the task involving $D_S$ when the BNA is GoogleNet~\cite{SzegedyCVPR15}, we use a batch size of 32 and a fixed learning rate $10^{-5}$/$10^{-6}$/$10^{-3}$ for step 1/2/3. The learning rate is chosen such that the trained network can converge under a reasonable amount of time. We set the weight of the softmax loss and the L2 loss in step 2 to be $10^3$ and 1 respectively such that both losses have comparable numerical values. Step 1/2/3 are trained for $10^4$/$10^3$/$10^3$ iterations. For the other training parameters, we adopt the default ones used in training the BVLC GoogleNet model~\cite{BVLCGoogleNetModel} for the ImageNet~\cite{DengCVPR09} classification task unless otherwise specified. In general, we adopt the default training parameters used in training each BNA for either the MNIST~\cite{LeCunIEEE98} or ImageNet~\cite{DengCVPR09} classification tasks in the Caffe~\cite{JiaCoRR14} and SqueezeNet\_v1.1~\cite{IandolaCoRR16} implementation unless otherwise specified.

%The details of the training parameters for all the other experiments are in the supplementary material.

\subsection{Performance References and Baselines}

To obtain the performance references of the fully supervised methods, we train a classifier with the BNA in Table~\ref{NN_architecture} in each domain using the T-R training data and labels in that domain. When the BNA is LeNet~\cite{LeNetPrototxt}, we train the classifier from scratch. For the other BNAs, we pre-train the classifier in the same way as that described in Sec.~\ref{subsec:training_details}. After training, for each DA task, we get two fully supervised classifiers $C_{fs,s}$ and $C_{fs,t}$ in the source and target domains respectively. For the baseline of the DA task, we directly feed the target-domain testing images to $C_{fs,s}$ to obtain the performance without applying any DA method. For the baseline of sensor fusion, we compare ZDDA$_3$ with a naive fusion method by predicting the label with the highest probability from $C_{RGB}$ and $C_D$ in Sec.~\ref{sec:method}.

\section{Experimental Result}

\begin{table}[t]
\fontsize{8}{9.6} \selectfont
\centering
\caption{\fontsize{9}{10.8} \selectfont The overall / average per class accuracy (\%) of the domain adaptation tasks (gray scale images $\rightarrow$ RGB images) formed by the datasets in Table~\ref{dataset_stats}, where we introduce the dataset IDs and use them to refer to the datasets here. The middle four rows show the performance of ZDDA$_2$. The color of each cell reflects the performance ranking in each column, where darker is better. The number in the parenthesis of the middle four rows is the semantic similarity between the T-R and T-I datasets measured by word2vec~\cite{w2v}, where larger numbers represent higher semantic similarity. The T-R target-domain training data is only available for the row ``target only"
}
\begin{tabular}{ccccc}
\\
\toprule
T-I &$D_M\rightarrow D_M$-M &$D_F\rightarrow D_F$-M &$D_N\rightarrow D_N$-M &$D_E\rightarrow D_E$-M\\
data &\includegraphics[width=0.075\linewidth]{mnist_sample.png} \raisebox{0.5mm}{$\rightarrow$} \includegraphics[width=0.075\linewidth]{mnist-m_sample.png}
&\includegraphics[width=0.075\linewidth]{fashion-mnist_sample.png} \raisebox{0.5mm}{$\rightarrow$} \includegraphics[width=0.075\linewidth]{fashion-mnist-m_sample.png} &\includegraphics[width=0.075\linewidth]{nist_sample.png} \raisebox{0.5mm}{$\rightarrow$} \includegraphics[width=0.075\linewidth]{nist-m_sample.png} &\includegraphics[width=0.075\linewidth]{emnist_sample.png} \raisebox{0.5mm}{$\rightarrow$} \includegraphics[width=0.075\linewidth]{emnist-m_sample.png}\\
\midrule[0.75pt]
source only &\cellcolor{blue!10}39.04/39.31 &\cellcolor{blue!10}33.77/33.77 &\cellcolor{blue!10}8.59/8.79 &\cellcolor{blue!10}33.70/33.70\\
\midrule
$D_M, D_M$-M &N/A &\cellcolor{blue!50}51.55/51.55 (0.049) &\cellcolor{blue!50}34.25/33.35 (0.174) &\cellcolor{blue!50}71.20/71.20 (0.178)\\
$D_F, D_F$-M &\cellcolor{blue!30}73.15/72.96 (0.049) &N/A &\cellcolor{blue!30}21.93/21.24 (0.059) &\cellcolor{blue!30}46.93/46.93 (0.053)\\
$D_N, D_N$-M &\cellcolor{blue!50}91.99/92.00 (0.174) &\cellcolor{blue!30}43.87/43.87 (0.059) &N/A &N/A\\
$D_E, D_E$-M &\cellcolor{blue!70}\color{white}94.84/94.82 (0.178) &\cellcolor{blue!70}\color{white}65.30/65.30 (0.053) &N/A &N/A\\
\midrule
target only &\cellcolor{blue!90}\color{white}97.33/97.34 &\cellcolor{blue!90}\color{white}84.44/84.44 &\cellcolor{blue!90}\color{white}62.13/61.99 &\cellcolor{blue!90}\color{white}89.52/89.52\\
\bottomrule
\end{tabular}
\label{exp_mnist-m}
\end{table}

We first compare ZDDA$_2$ with the baseline in four domain adaptation (DA) tasks (adapting from gray scale to RGB images) involving $D_M$, $D_F$, $D_N$, and $D_E$, and the result is summarized in Table~\ref{exp_mnist-m}, where the first two numbers represent the overall/average per class accuracy (\%). Darker cells in each column represent better classification accuracy in each task. In Table~\ref{exp_mnist-m}, the middle four rows represent the performance of ZDDA$_2$. \{$D_N$, $D_N$-M\} and \{$D_E$, $D_E$-M\} cannot be the T-I data for each other because they are both directly related to the letter classification tasks. Table~\ref{exp_mnist-m} shows that regardless of which T-I data we use, ZDDA$_2$ significantly outperforms the baseline (source only). To see how the semantic similarity between the T-R dataset (denoted as $D_{T-R}$) and T-I dataset (denoted as $D_{T-I}$) affects the performance, we are inspired by~\cite{FuCVPR15} and use the word2vec~\cite{w2v} to compute the mean similarity (denoted as $S$) of any two labels from $D_{T-R}$ and $D_{T-I}$ (one from each). We report $S$($D_{T-R}$, $D_{T-I}$) in the parenthesis of the middle four rows of Table~\ref{exp_mnist-m}, where higher $S$ represents higher semantic similarity. Given Table~\ref{exp_mnist-m} and the following reference $S$ values: $S$(object, scene)=0.192, $S$(animal, fruit)=0.171, and $S$(cat, dog)=0.761, we find that: (1) For all the listed DA tasks except $D_F\rightarrow D_F$-M, higher $S$ corresponds to better performance, which is consistent with our intuition that using more relevant data as the T-I data improves the performance more. (2) All the listed $S$s in Table~\ref{exp_mnist-m} are close to or lower than $S$(animal, fruit)=0.171, which we believe shows that our T-I data is highly irrelevant to the T-R data.

\begin{table}[t]
\centering
\caption{\fontsize{9}{10.8} \selectfont The performance comparison of the domain adaptation task MNIST$\rightarrow$MNIST-M. The color of each cell reflects the performance ranking (darker is better). For ZDDA$_2$, we report the best overall accuracy from Table~\ref{exp_mnist-m}. \textbf{All the listed methods except ZDDA$_2$ use the MNIST-M training data.} Without the access to the MNIST-M training data, ZDDA$_2$ can still achieve the accuracy comparable to those of the competing methods (even outperform most of them) in this task
}
\begin{tabular}{ccccccc}
\\
\toprule
method &\cite{GaninJMLR16} &\cite{SenerNIPS16} &\cite{HaeusserICCV17} &\cite{SaitoICML17} &\cite{BousmalisCVPR17} &ZDDA$_2$\\
\midrule
accuracy (\%) &\cellcolor{blue!10}76.66 &\cellcolor{blue!20}86.70 &\cellcolor{blue!30}89.53 &\cellcolor{blue!40}94.20 &\cellcolor{blue!60}\color{white}98.20 &\cellcolor{blue!50}94.84\\
\bottomrule
\end{tabular}
\label{exp_mnist-m_sota}
\end{table}

\begin{table}[t]
\centering
\fontsize{8}{9.6} \selectfont
\caption{\fontsize{9}{10.8} \selectfont Performance comparison with different numbers of classes in scene classification. The reported numbers are classification accuracy (\%). The color of each cell reflects the performance ranking in each column, where darker color means better performance. $P_\text{RGB-D}$ represents the \textbf{task-irrelevant} RGB-D pairs}
\begin{tabular}{ccccccccccccc}
\\
\toprule
exp. & &training &testing &\multicolumn{9}{c}{number of classes}\\
ID &method &modality &modality &2 &3 &4 &5 &6 &7 &8 &9 &10\\
\midrule
1 &GoogleNet &D &D &\cellcolor{blue!20}85.53 &\cellcolor{blue!20}83.33 &\cellcolor{blue!20}82.89 &\cellcolor{blue!20}70.00 &\cellcolor{blue!20}67.11 &\cellcolor{blue!20}59.02 &\cellcolor{blue!20}54.28 &\cellcolor{blue!20}50.88 &\cellcolor{blue!20}51.84\\
2 &ZDDA$_2$ &D+$P_\text{RGB-D}$ &D &\cellcolor{blue!60}\color{white}88.16 &\cellcolor{blue!50}85.96 &\cellcolor{blue!30}83.55 &\cellcolor{blue!50}77.89 &\cellcolor{blue!30}70.18 &\cellcolor{blue!40}66.92 &\cellcolor{blue!40}64.80 &\cellcolor{blue!30}62.28 &\cellcolor{blue!30}59.74\\
3 &ZDDA$_3$ &D+$P_\text{RGB-D}$ &D &\cellcolor{blue!60}\color{white}88.16 &\cellcolor{blue!60}\color{white}86.84 &\cellcolor{blue!50}84.87 &\cellcolor{blue!50}77.89 &\cellcolor{blue!40}72.37 &\cellcolor{blue!40}66.92 &\cellcolor{blue!30}64.47 &\cellcolor{blue!50}64.33 &\cellcolor{blue!50}63.16\\
\midrule
4 &GoogleNet &D &RGB &68.42 &57.02 &56.58 &48.95 &42.11 &45.11 &40.46 &34.50 &31.58\\
5 &ZDDA$_2$ &D+$P_\text{RGB-D}$ &RGB &\cellcolor{blue!10}80.26 &\cellcolor{blue!10}78.07 &\cellcolor{blue!10}76.32 &\cellcolor{blue!10}67.37 &\cellcolor{blue!10}57.89 &\cellcolor{blue!10}53.76 &\cellcolor{blue!10}47.37 &\cellcolor{blue!10}45.03 &\cellcolor{blue!10}43.16\\
6 &GoogleNet &RGB &RGB &\cellcolor{blue!60}\color{white}88.16 &\cellcolor{blue!30}85.09 &\cellcolor{blue!50}84.87 &\cellcolor{blue!60}\color{white}79.47 &\cellcolor{blue!60}\color{white}78.07 &\cellcolor{blue!60}\color{white}68.80 &\cellcolor{blue!60}\color{white}70.07 &\cellcolor{blue!60}\color{white}69.88 &\cellcolor{blue!60}\color{white}63.68\\
\midrule
7 &ZDDA$_3$ &D+$P_\text{RGB-D}$ &RGB-D &\cellcolor{blue!60}\color{white}88.16 &\cellcolor{blue!50}85.96 &\cellcolor{blue!60}\color{white}85.53 &\cellcolor{blue!30}76.32 &\cellcolor{blue!50}72.81 &\cellcolor{blue!50}68.42 &\cellcolor{blue!50}65.13 &\cellcolor{blue!40}63.16 &\cellcolor{blue!50}63.16\\
\midrule\multicolumn{4}{c}{selected scene IDs (defined in Sec.~\ref{subsec:dataset})} &0$\sim$1 &0$\sim$2 &0$\sim$3 &0$\sim$4 &0$\sim$5 &0$\sim$6 &0$\sim$7 &0$\sim$8 &0$\sim$9\\
\bottomrule
\end{tabular}
\label{result_different_number_of_classes}
\end{table}

Second, in Table~\ref{exp_mnist-m_sota}, we compare ZDDA$_2$ with the existing DA methods because the DA task $D_M\rightarrow D_M$-M is considered as one of the standard experiments in recent works~\cite{BousmalisCVPR17,GaninJMLR16,HaeusserICCV17,SaitoICML17,SenerNIPS16}. Although this is not a fair comparison (because ZDDA$_2$ has no access to the T-R target-domain training data), we find that ZDDA$_2$ can reach the accuracy comparable to those of the state-of-the-art methods (even outperform some of them), which supports that ZDDA$_2$ is a promising DA method when the T-R target-domain training data is unavailable.

Third, we test the efficacy of ZDDA on the DA tasks constructed from $D_S$ (adapting from depth to RGB images). We compare ZDDA with the baseline under different scene classification tasks by changing the number of scenes involved. The result is summarized in Table~\ref{result_different_number_of_classes}, where we list the training and testing modalities for each method. We also list the scene IDs (introduced in Sec.~\ref{subsec:dataset}) involved in each task. Darker cells represent better accuracy in each column. We verify the irrelevance degree between T-R and T-I data by measuring the semantic similarity using the word2vec~\cite{w2v} (the same method we use in Table~\ref{exp_mnist-m}). For the 10-class experiment in Table~\ref{result_different_number_of_classes}, $S$($D_S$(T-R), $D_S$(T-I))=0.198 (close to the reference $S$(object, scene)=0.192), which we believe shows high irrelevance between our T-I and T-R data. For simplicity, we use $E_i$ to refer to the experiment specified by exp. ID $i$ in this section. For the fully supervised methods in depth domain, ZDDA ($E_2$, $E_3$) outperforms the baseline ($E_1$) due to the extra information brought by the T-I RGB-D pairs. We find that for most listed tasks, ZDDA$_3$ ($E_3$) outperforms ZDDA$_2$ ($E_2$), which is consistent with our intuition because the source representation in ZDDA$_2$ is constrained by the L2 loss, while the counterpart in ZDDA$_3$ is learned without the L2 constraint given the simulated target representation. The fully supervised method in RGB domain ($E_6$) outperforms the baseline of the domain adaptation ($E_4$) and ZDDA$_2$ ($E_5$) because $E_6$ has access to the T-R RGB training data which is unavailable for $E_4$ and $E_5$. The performance improvement from $E_4$ to $E_5$ is caused by ZDDA$_2$'s training procedure as well as the extra T-I RGB-D training pairs. $E_3$ and $E_7$ perform similarly, which supports that the simulated target representation in ZDDA$_3$ is similar to the real one.

\begin{table}[t]
\centering
\caption{\fontsize{9}{10.8} \selectfont Validation of ZDDA's performance (in mean classification accuracy (\%)) with different training/testing splits and choices of classes in scene classification. GN stands for GoogleNet~\cite{SzegedyCVPR15}. The definition of $P_\text{RGB-D}$ and the representation of the cell color in each column are the same as those in Table~\ref{result_different_number_of_classes}
}
\begin{tabular}{ccccc}
\\
\toprule
 &training &testing &validation on &validation on\\
method &modality &modality &train/test splits &class choices\\
\midrule
GN &D &D &\cellcolor{blue!20}52.63$\pm$1.76 &\cellcolor{blue!20}53.98$\pm$1.68\\
ZDDA$_2$ &D+$P_\text{RGB-D}$ &D &\cellcolor{blue!30}56.89$\pm$2.13 &\cellcolor{blue!30}62.05$\pm$1.97\\
ZDDA$_3$ &D+$P_\text{RGB-D}$ &D &\cellcolor{blue!40}58.37$\pm$3.08 &\cellcolor{blue!50}62.49$\pm$1.74\\
\midrule
GN &D &RGB &31.26$\pm$1.76 &32.60$\pm$2.37\\
ZDDA$_2$ &D+$P_\text{RGB-D}$ &RGB &\cellcolor{blue!10}44.47$\pm$2.50 &\cellcolor{blue!10}45.47$\pm$2.57\\
GN &RGB &RGB &\cellcolor{blue!60}\color{white}66.26$\pm$1.60 &\cellcolor{blue!60}\color{white}67.95$\pm$2.20\\
\midrule
ZDDA$_3$ &D+$P_\text{RGB-D}$ &RGB-D &\cellcolor{blue!50}58.68$\pm$3.10 &\cellcolor{blue!40}62.13$\pm$1.50\\
\midrule
\multicolumn{3}{c}{\# of classes / \# of folds} &10 / 5 &9 / 10\\
\bottomrule
\end{tabular}
\label{cv_split_class_selection}
\end{table}

\begin{table}[t]
\centering
\caption{\fontsize{9}{10.8} \selectfont Validation of ZDDA's performance with different base network architectures in scene classification. The reported numbers are classification accuracy (\%). The definition of $P_\text{RGB-D}$ and the representation of the cell color in each column are the same as those in Table~\ref{result_different_number_of_classes}
}
\begin{tabular}{cccccc}
\\
\toprule
 &training &testing &\multicolumn{3}{c}{base network architecture}\\
method &modality &modality &GoogleNet~\cite{SzegedyCVPR15} &AlexNet~\cite{KrizhevskyNIPS12} &SqueezeNet\_v1.1~\cite{IandolaCoRR16}\\
\midrule
BNA &D &D &\cellcolor{blue!20}51.84 &\cellcolor{blue!20}49.74 &\cellcolor{blue!20}48.68\\
ZDDA$_2$ &D+$P_\text{RGB-D}$ &D &\cellcolor{blue!30}59.74 &\cellcolor{blue!40}51.05 &\cellcolor{blue!50}56.32\\
ZDDA$_3$ &D+$P_\text{RGB-D}$ &D &\cellcolor{blue!50}63.16 &\cellcolor{blue!40}51.05 &\cellcolor{blue!50}56.32\\
\midrule
BNA &D &RGB &31.58 &30.26 &26.58\\
ZDDA$_2$ &D+$P_\text{RGB-D}$ &RGB &\cellcolor{blue!10}43.16 &\cellcolor{blue!10}40.00 &\cellcolor{blue!10}35.79\\
BNA &RGB &RGB &\cellcolor{blue!60}\color{white}63.68 &\cellcolor{blue!60}\color{white}59.47 &\cellcolor{blue!60}\color{white}57.37\\
\midrule
ZDDA$_3$ &D+$P_\text{RGB-D}$ &RGB-D &\cellcolor{blue!50}63.16 &\cellcolor{blue!50}51.84 &\cellcolor{blue!30}56.05\\
\bottomrule
\end{tabular}
\label{cv_NN_architecture}
\end{table}

To test the consistency of the performance of ZDDA compared to that of the baseline, we perform the following three experiments. First, we conduct 5-fold cross validation with different training/testing splits for the 10-scene classification. Second, we perform 10-fold validation with different selections of classes for the 9-scene classification (leave-one-class-out experiment out of the 10 selected scenes introduced in Sec.~\ref{subsec:dataset}). Third, we validate ZDDA's performance with different base network architectures. The results of the first two experiments are presented in Table~\ref{cv_split_class_selection}, and the result of the third experiment is shown in Table~\ref{cv_NN_architecture}. The results of Table~\ref{result_different_number_of_classes}, Table~\ref{cv_split_class_selection}, and Table~\ref{cv_NN_architecture} are consistent.

%The results of Table~\ref{cv_split_class_selection} and Table~\ref{cv_NN_architecture} are consistent with our observations from Table~\ref{result_different_number_of_classes}.

%The results of the first two experiments are presented in Table~\ref{cv_split_class_selection} (the extended version is in the supplementary material), and the result of the third experiment is shown in Table~\ref{cv_NN_architecture}.

\begin{figure}[t]
    \centering
    \subfloat[\small naive fusion]{%
        \includegraphics[width=0.32\linewidth]{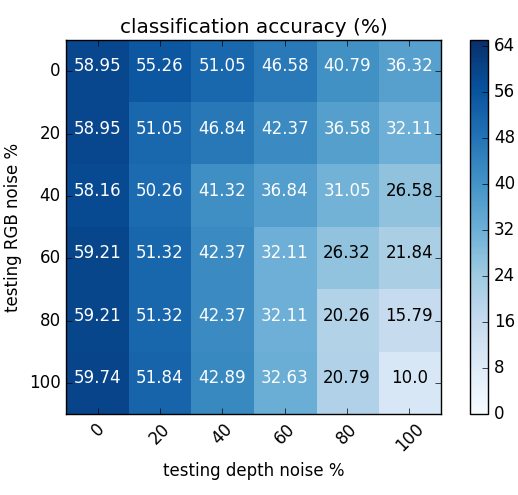}%
        \label{fusion:a}%
        }%
    \hfill%
    \subfloat[\small ZDDA$_3$]{%
        \includegraphics[width=0.32\linewidth]{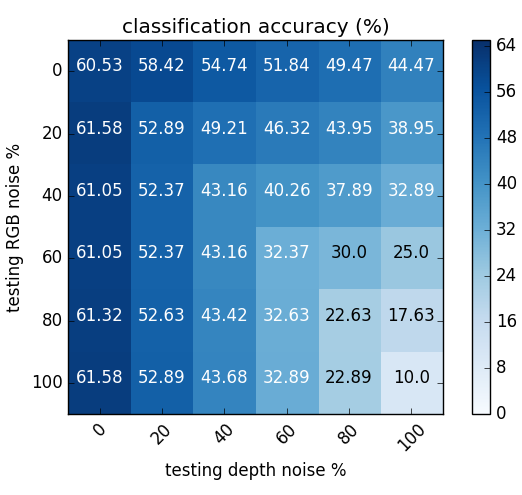}%
        \label{fusion:b}%
        }%
    \hfill%
    \subfloat[\small accuracy diff. ((b)-(a))]{%
        %accuracy improvement from (a) to (b)
        \includegraphics[width=0.35\linewidth]{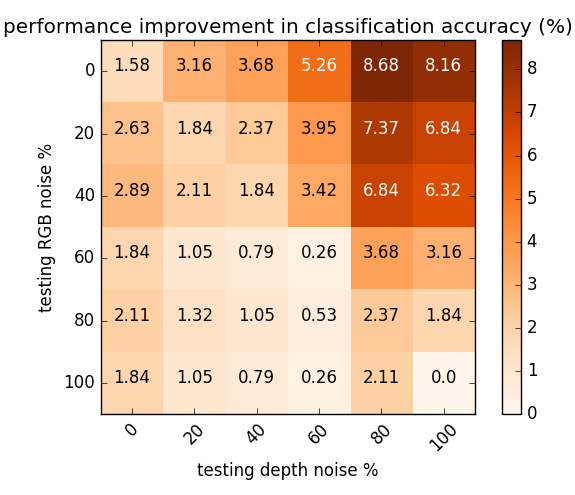}%
        \label{fusion:c}%
        }%
    \caption{\fontsize{9}{10.8} \selectfont Performance comparison between the two sensor fusion methods with black images as the noisy images. We compare the classification accuracy (\%) of (a) naive fusion and (b) ZDDA$_3$ under different noise levels in both RGB and depth testing data. (c) shows that ZDDA$_3$ outperforms the naive fusion under most conditions}
    \label{fusion}
\end{figure}

In Table~\ref{result_different_number_of_classes}, Table~\ref{cv_split_class_selection}, and Table~\ref{cv_NN_architecture}, the classification accuracy is reported under the condition of noise-free training and testing data. To let ZDDA be more robust to noisy input, we train ZDDA$_3$ step 3 with noisy training data (we use $p_{train}=20$ as explained in Sec.~\ref{subsec:training_details}), and evaluate the classification accuracy under different noise conditions for both RGB and depth testing data. The result is presented in Fig.~\ref{fusion}, where ZDDA$_3$ (Fig.~\ref{fusion:b}) outperforms the naive fusion method (Fig.~\ref{fusion:a}) under most conditions, and the performance improvement is shown in Fig.~\ref{fusion:c}. Both Fig.~\ref{fusion:a} and Fig.~\ref{fusion:b} show that the performance degradation caused by the noisy depth testing data is larger than that caused by the noisy RGB testing data, which supports that the trained RGB-D classifier relies more on the depth domain. Traditionally, training a fusion model requires the T-R training data in both modalities. However, we show that without the T-R training data in the RGB domain, we can still train an RGB-D fusion model, and that the performance degrades smoothly when the noise increases. In addition to using black images as the noise model, we evaluate the same trained joint classifier in ZDDA$_3$ using another noise model (adding a black rectangle with a random location and size to the clean image) at testing time, and the result also supports that ZDDA$_3$ outperforms the naive fusion method. Although we only use black images as the noise model for ZDDA$_3$ at training time, we expect that adding different noise models can improve the robustness of ZDDA$_3$.

\section{Conclusion and Future Work}

We propose zero-shot deep domain adaptation (ZDDA), a novel approach to perform domain adaptation (DA) and sensor fusion with no need of the task-relevant target-domain training data which can be inaccessible in reality. Rather than solving the zero-shot DA problem in general, we aim at solving the problems under the assumption that task-relevant source-domain data and task-irrelevant dual-domain paired data are available. Our key idea is to use the task-relevant source-domain data to simulate the task-relevant target-domain representations by learning from the task-irrelevant dual-domain pairs. Experimenting on the MNIST~\cite{LeCunIEEE98}, Fashion-MNIST~\cite{XiaoCoRR17}, NIST~\cite{Grother16}, EMNIST~\cite{CohenCoRR17}, and SUN RGB-D~\cite{SongCVPR15} datasets, we show that ZDDA outperforms the baselines in DA and sensor fusion even without the task-relevant target-domain training data. In the task adapting from MNIST~\cite{LeCunIEEE98} to MNIST-M~\cite{GaninICML15}, ZDDA can even outperform several state-of-the-art DA methods which require access to the MNIST-M~\cite{GaninICML15} training data. One industrial use case which we plan to apply ZDDA to in our follow-up work is training an RGD object classifier given only the textureless CAD models of those objects. In this case, depth and RGB images are source and target domains, respectively. The depth images can be rendered from the provided CAD models, and publicly available RGB-D datasets can serve as the task-irrelevant RGB-D data. We believe that ZDDA can be straightforwardly extended to handle other tasks of interest by modifying the loss functions in Fig.~\ref{training_flow} step 2 and step 3.

%, and we share a template of ZDDA's training procedure and two examples of such extensions in the supplementary material.

%\bibliographystyle{splncs04}
%\bibliography{egbib}

\end{document}